\documentclass[10pt,twocolumn,letterpaper]{article}
\usepackage[pagenumbers]{cvpr} % To force page numbers, e.g. for an arXiv version

\usepackage{graphicx}
\usepackage{amsmath}
\usepackage{amssymb}
\usepackage{booktabs}
\usepackage[colorlinks=true, allcolors=blue]{hyperref}
\usepackage{comment}
\usepackage{afterpage}

\usepackage[capitalize]{cleveref}
\crefname{section}{Sec.}{Secs.}
\Crefname{section}{Section}{Sections}
\Crefname{table}{Table}{Tables}
\crefname{table}{Tab.}{Tabs.}

\def\cvprPaperID{*****} % *** Enter the CVPR Paper ID here
\def\confName{CVPR}
\def\confYear{2022}
\makeatletter
\newenvironment{tablehere}
{\def\@captype{table}}
{
	
}
\newenvironment{figurehere}
{\def\@captype{figure}}
{}
\makeatother
\newcommand{\bunderline}[2][4]{\underline{#2\mkern-#1mu}\mkern#1mu}

\newcommand{\nunder}[2][5]{\mathrlap{\mkern\the\numexpr#1/2mu\relax\underline{\phantom{\mathrm{#2}\mkern-#1mu}}}#2}

\usepackage{amsmath}
\makeatletter
\newsavebox\tboxa
\newsavebox\tboxb
\newlength\tdima

\newcommand*{\oversymb}{\mathpalette\@oversymb}

\newcommand*{\@oversymb}[2]{%
    \sbox{\tboxa}{$\m@th#1\mathrm{#2}$}%
    \setbox\tboxb\null%
    \ht\tboxb\ht\tboxa%
    \dp\tboxb\dp\tboxa%
    \wd\tboxb\wd\tboxa%
    \sbox{\tboxa}{$\m@th#1{#2}$}%
    \setlength\tdima{\the\wd\tboxa}%
    \addtolength\tdima{-\the\wd\tboxb}%
    \sbox{\tboxb}{$\m@th#1\hskip\tdima\overline{\xusebox{\tboxb}}$}%
    \rlap{\usebox\tboxb}{\usebox\tboxa}}

\newcommand*{\xusebox}[1]{\mathord{{\usebox{#1}}}}

\makeatother
\usepackage{subcaption}
\usepackage{stackengine}
\usepackage{mathtools}% Loads amsmath

\begin{document}

\title{On-board Sonar Data Classification for Path Following in Underwater Vehicles using Fast Interval Type-2 Fuzzy Extreme Learning Machine }

\author{Adrian Rubio-Solis$^{1}$,  Luciano Nava-Balanzar$^{2}$ and Tomas Salgado-Jimenez$^{2}$
\\
\textit{$^{1}$Hamlyn Centre for Robotic
Surgery, Imperial College London, London, SW7 2AZ, UK} \\
\textit{$^{2}$Center for Engineering and Industrial Development (CIDESI), Qro. M\'{e}xico.} \\
{\tt\small arubioso@ic.ac.uk, Inava@cidesi.edu.mx, t.salgado@cidesi.edu.mx}
% \and
}
\maketitle

%%%%%%%%% ABSTRACT
\begin{abstract}
In autonomous underwater missions, the successful completion of predefined paths mainly depends on the ability of underwater vehicles to recognise their surroundings. In this study, we apply the concept of Fast Interval Type-2 Fuzzy Extreme Learning Machine (FIT2-FELM) to train a Takagi-Sugeno-Kang IT2 Fuzzy Inference System (TSK IT2-FIS) for on-board sonar data classification using an underwater vehicle called BlueROV2. The TSK IT2-FIS is integrated into a Hierarchical Navigation Strategy (HNS) as the main navigation engine to infer local motions and provide the BlueROV2 with full autonomy to follow an obstacle-free trajectory in a water container of $2.5m \times 2.5m \times 3.5m$. Compared to traditional navigation architectures, using the proposed method, we observe a robust path following behaviour in the presence of uncertainty and noise. We found that the proposed approach provides the BlueROV with a more complete sensory picture about its surroundings while real-time navigation planning is performed by the concurrent execution of two or more tasks.
\end{abstract}

%%%%%%%%% BODY TEXT
\section{Introduction}
\label{sec:intro}
Path following has become one of the most frequent underwater missions that usually involves the inspection of underwater structures such as dams, ship hulls, harbors and oil pipelines \cite{xanthidis2020navigation,panda2020comprehensive,jacobi2015autonomous,liang2017three}. Other submarine missions that involve completing a predefined path include mapping and monitoring of the marine environment and data collection in the deep ocean \cite{qu2020fuzzy, xiang2017robust}. It is known that in submarine applications the use of fuzzy logic usually provides a better treatment of nonlinearities of the underwater vehicle dynamics, intrinsic uncertainties in submarine environments and noisy sensor readings\cite{ xiang2017robust, liang2016path,liang2018three}. 

 In this study, an initial path following approach for underwater vehicles using Fast Interval Type-2 Fuzzy Extreme Learning Machine (FIT2-FELM) is presented. 
%///////////////////////////////////////
\begin{figurehere}
\begin{center}
%\centering
%\vspace{2mm}
\includegraphics[width=8cm, height=6cm ]{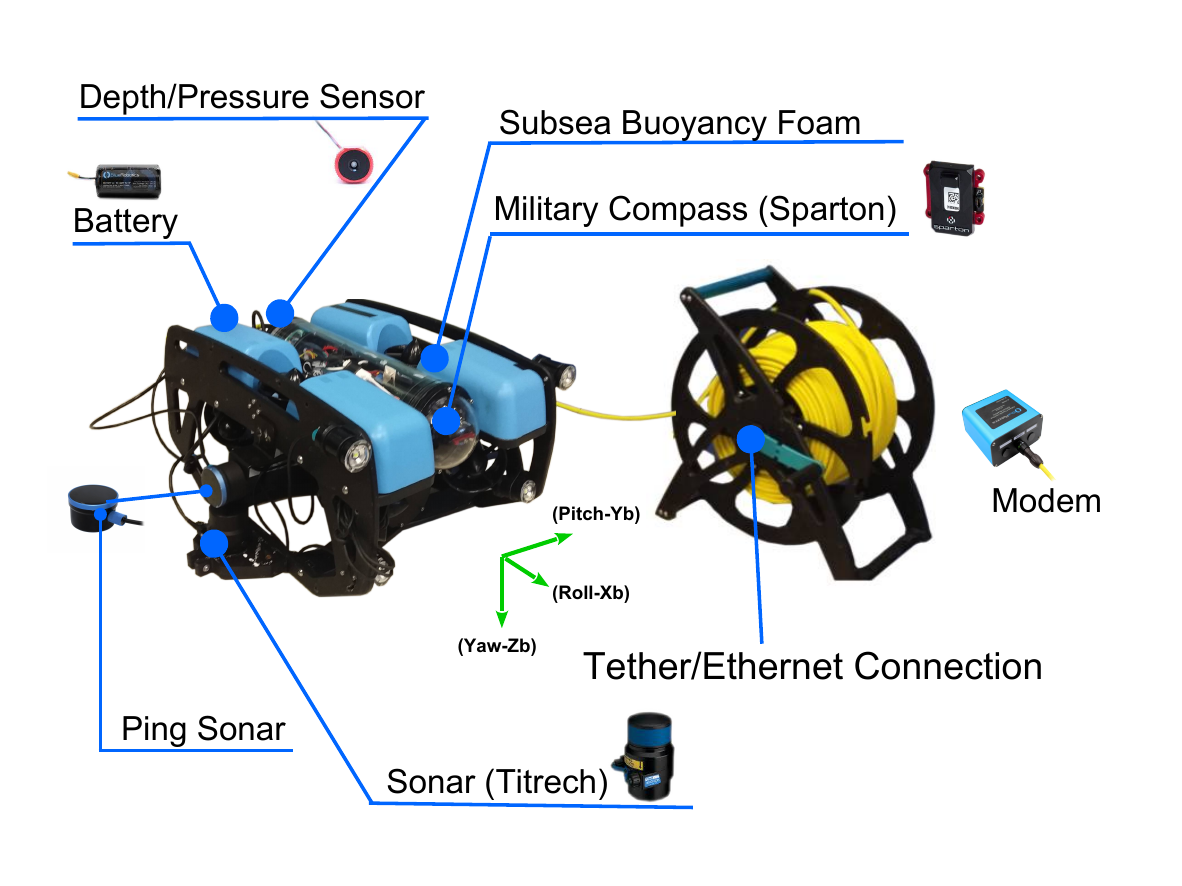}
\caption{\footnotesize Underwater Vehicle BlueROV2 and its sensory system integrated at CIDESI, Mexico.}\label{fig::ROV_standalone}
\end{center}
\end{figurehere}
The proposed FIT2-FELM is a fast learning method to the training of TSK Interval Type-2 Fuzzy Inference Systems (FIS). The TSK IT2-FIS is incorporated to a Hierarchical Navigation Strategy (HNS) for on-board classification of sonar data and used by an underwater vehicle called BlueROV2 to recognise its surroundings. 
%...............................................
\begin{figure*}[t!]
\begin{center}
%\centering
%\vspace{2mm}
\includegraphics[width=17.1cm, height=6.0cm ]{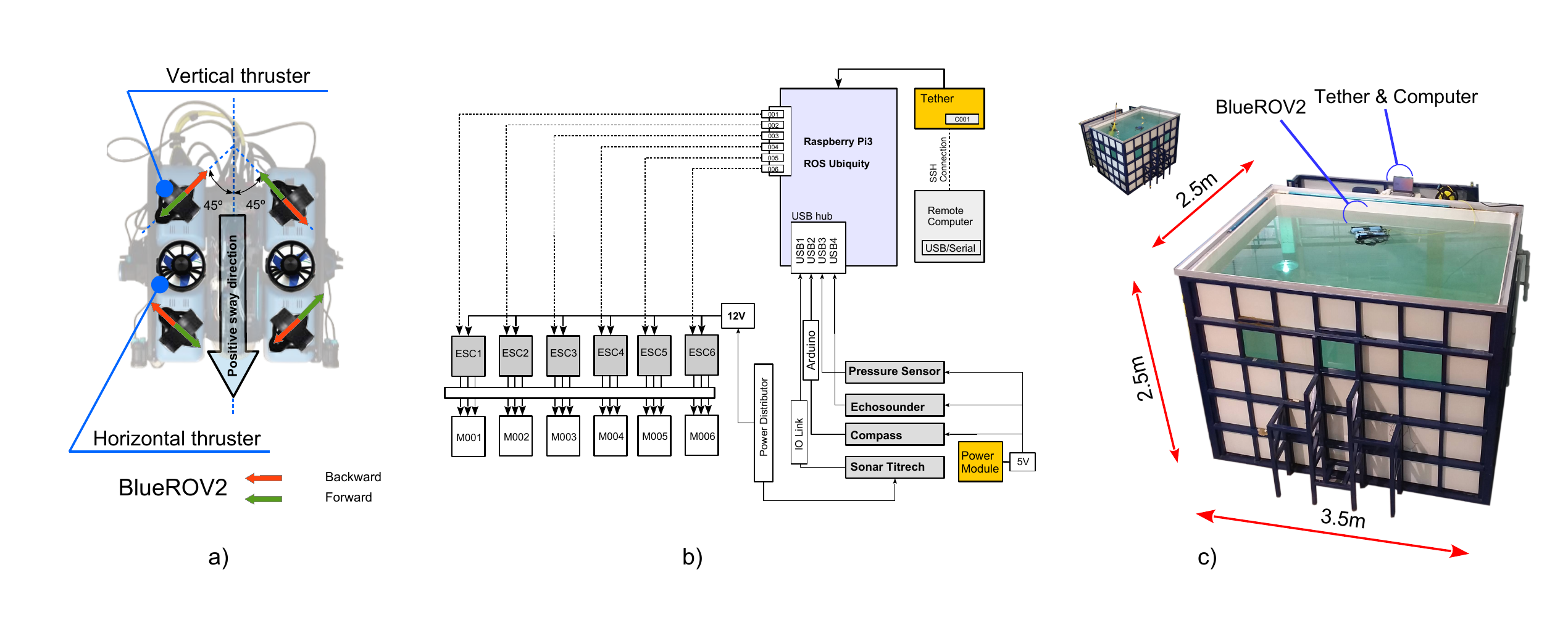}
\caption{\footnotesize (a) Thruster Configuration of BlueROV2, (b) System's configuration used by the BlueROV2 and (c) Water container used for experiments.}\label{fig::ROV_components}
\end{center}
\end{figure*}
The HNS employs this information as a navigation mechanism to infer local control behaviours while providing the BlueROV2 with full autonomy to navigate predefined obstacle-free circuits in a water container of $2.5m \times 2.5m \times \ 3.5m$ (Fig. \ref{fig::ROV_standalone}, \ref{fig::ROV_components}). In this work, 
the path following task used in all experiments is a common application of remotely operated vehicles (ROVs) for the inspection of underwater structures \cite{jacobi2015autonomous,liang2017three}. The predefined mission involves following a trajectory (circuit) that is defined by the contour of the water container at different depths. To complete one path, the HNS follows a bottom-up hierarchy with an increasing behavioral complexity that consists of two control levels. At low level, for local planning, the HNS utilises the classification information provided by the IT2-FIS to break down the navigation strategy into a number of four fuzzy behaviours. The first three behaviours are realised by implementing an IT2 Fuzzy Proportional Derivative Control (IT2-FPDC) to control vehicle's heading, its vertical position (depth) and its distance to each wall of the container. The last behaviour called contour classification is performed by the IT2-FIS in order to discriminate signals collected by an onboard sonar. Each behaviour consists of a nonlinear mapping from different subsets of the BlueROV2's sensor suite to a common actuator by operating concurrently (Fig. \ref{fig::ROV_standalone}). At the high level, a number of IF-THEN rules is implemented to coordinate all local controllers as a set of more complex behaviours. 

% all fuzzy behaviours are organised by a number of IF-THEN rules as a set of building blocks for more intelligent composite behaviours. 

Using FIT2-FELM, the HNS provides the BlueROV2 with a complete sensory picture about its surroundings. Thus, path following is decomposed into behavioural functions for wall-following, local path planning and collision avoidance. From our experiments, the BlueROV2 reached an accuracy for sonar data classification of $92.7\%$ in real time conditions. Such experiments highlight the relevance of accurate sonar data classification to local planning while facilitating an accurate path following behaviour in indoor environments.

\section{Methods}
\subsection{Experiments with the BlueROV2}
The BlueROV2 is an underwater vehicle equipped  with a six-thruster vectored configuration as illustrated in Fig. \ref{fig::ROV_components}(a). Two out of six thrusters are oriented horizontally while the remaining are oriented in a vertical direction. This configuration provides the underwater vehicle with the ability to move itself up-down as well as to control its yaw orientation and moves forward and backward. As shown in Fig. \ref{fig::ROV_standalone}, four subsea buoyancy foams are mounted to the BlueROV2's structure in order to induce a righting moment that results in a vertical motion largely decoupled from the lateral motion. 
%Because of the righting moment induced by four subsea buoyancy foams (see Fig. \ref{fig::ROV_standalone}), the BlueROV2's dynamics are such that the vertical motion is largely decoupled from the lateral motion.

Experiments test the capability of the BlueROV2 to autonomously complete a predefined circuit in underwater structures. To enable the BlueROV2 to detect and recognise its surroundings, an open-source electronics that consists of a 1) pressure sensor, 2) ping sonar, 3) data sonar, 4) compass and a 5) Raspberri Pi3 was integrated (see Fig. \ref{fig::ROV_components}(b)). The pressure sensor is able to measure up to 30 Bar (300m depth) with a depth resolution of 2mm, while 2) the ping sonar is able to measure distances up to 30 meters. The data sonar has a range resolution of $7.5$mm, a beam width of $3^{\circ}$, and a variable scanned sector as shown in Fig. \ref{fig::ROV_components}(b). The compass is a micro-sized and light weight attitude heading sensor with a static heading accuracy of $0.2^{\circ}$ RMS and full $360^{\circ}$ rollover capability. The main computer used by the BlueROV2 is a Raspberri Pi3, in which the middleware Robotic Operating System (ROS ubiquity) was installed to implement all machine learning algorithms and controllers. A line SSH connection between the Raspberri Pi3 and an Ubuntu computer was used to monitor all sensor values and define the parameters of each experiment (See Fig. \ref{fig::ROV_components}). An indoors water container whose dimensions are $2.5 \times 2.5 \times 3.5$ metres was employed to carry out all experiments as shown in Fig. \ref{fig::ROV_components}(c)). To reproduce some undersea conditions, salty water with a density of about $1028kg/m^3$ was added to the container. Sonar data was collected using a scanning sector of five angles (see Fig. \ref{fig::Piscina}), i.e. $[180^o, 172^o, 164^o, 156^o, 148^o]$ while the BlueROV2 is moving at constant speed of 30 mm/sec across the closest wall/corner. A low-pass filtered data set of $788 \times 5$ records was collected at different depth and locations.  
\subsection{Proposed FIT2-FELM}
FIT2-FELM is a fast learning strategy to train a class of IT2 Fuzzy Inference Systems (IT2-FISs) for the classification of on-board sonar signals. 
\begin{figurehere}
\begin{center}
%\centering
%\vspace{2mm}
\includegraphics[width=7.5cm, height=6cm ]{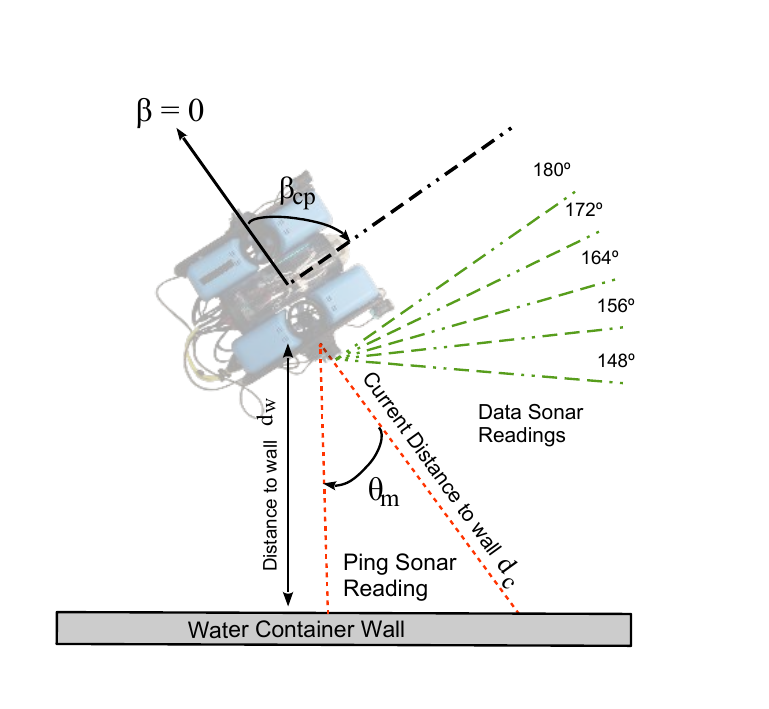}
\caption{\footnotesize Scanning sector configuration used by the on-board sonar for all experiments.}\label{fig::Piscina}
\end{center}
\end{figurehere}
%+++++++++++++++++++++++++++++++++++++++++++++++

For autonomous navigation, a Multi-Input-Single-Output (MISO) IT2-FIS is used as a leading information model to classify the surroundings of the BlueROV2. For $'P'$ different sonar training samples $(\textbf{x}_p, \textbf{t}_p)$, a MISO IT2-FIS is a neural structure with a given number of IT2 fuzzy rules $'M'$, in which, each $ith$ output is \cite{rubio2017ensemble}:  \cite{rubio2014interval}:
\begin{equation}
    y_p = \frac{1}{2}\left(y_l + y_r \right);~ p=1,\ldots, P
\end{equation}
%+++++++++++++++++++++++++++++++++++++++++++++++
where $\textbf{x}_p = \{x_{p1}, \ldots, x_{pN}\} \in \textbf{R}^{N}$ is an input vector, and $\textbf{t}_p = [t_{p1}, \ldots, t_{p\tilde{N}}]^T \in \textbf{R}^{\tilde{N}}$ the target.  The number of outputs in the FIT2-FELM is denoted by $N$. Here, a product inference rule base for an IT2-FIS  of Takagi-Sugeno-Kang (TSK) is considered, where each rule is given by \cite{rubio2018evolutionary}:
\begin{multline}
R^j: \text{IF}~x_{p1}~\text{is}~\tilde{A}_{j1} ~\text{AND}~ x_{p2}~\text{is}~\tilde{A}_{j2}~\text{AND}~ \ldots \\
\text{IF}~x_N~\text{is}~A_{jN}~\text{THEN}~w_{j} =  q_{j,1} x_1 + \ldots  q_{j,N} x_N
\end{multline}

in which, $k=1 \ldots, N, j = 1, \ldots, M$ and each $\tilde{A}_{jk}$ is an IT2 Fuzzy Set (FS) of the $kth$ input variable $x_k$ \cite{rubio2018evolutionary}. When $w_{j}$ is a crisp value, the FIS is of Mamdani type. From Eq. (2), an IT2 FS $\tilde{A}_{ik}$ uses a primary Gaussian MF with a variable width $[\sigma_{jk}^1, \sigma_{jk}^2]$ and a fixed mean $m_{jk}$. In IT2 Fuzzy Logic Systems (FLS), the Footprint Of Uncertainty (FOU) is represented by its lower and upper MF $[\bunderline[0]{ \mu}_{\tilde{A}_{jk} },\oversymb{\mu}_{\tilde{A}_{jk}} ]$ 
\begin{equation}
   [\bunderline[0]{ \mu}_{\tilde{A}_{jk} },\oversymb{\mu}_{\tilde{A}_{jk}} ]:=
    \begin{cases}
    \text{exp} \left[ -\frac{1}{2} \left( \frac{x_{pk} - m_{jk}}{\sigma_{jk}^1} \right)^2 \right] \vspace{2mm} \\
\text{exp} \left[ -\frac{1}{2} \left( \frac{x_{pk} - m_{jk}}{\sigma_{jk}^2} \right)^2 \right]
    \end{cases}
\end{equation} 
The firing strength $\tilde{F}^j$ of each $jth$ fuzzy rule is computed by performing fuzzy meet operation with the inputs using an algebraic product operation:
\begin{equation}
\tilde{F}^j = [\bunderline[3]{f}_j(\vec{x}_p),\oversymb{f}_j(\vec{x}_p)] 
\end{equation}
\vspace{3mm}
%++++++++++++++++++++++++++++++++++++++++++
\begin{tablehere}
\captionsetup{font=scriptsize,
              singlelinecheck = false}
\caption{\footnotesize SC algorithm for calculating end points $y_l$ and $y_r$.}\label{iris_results} % title of Table
%\vspace{0.0cm}
\centering % used for centering table
\begin{tabular}{p{0.7cm} p{1.5cm} p{1cm} }
\hline 
%inserts double horizontal lines
 %& \multicolumn{2}{c}{RMSE} \\
\multicolumn{1}{l}{\scriptsize Step}& \multicolumn{1}{c}{\scriptsize Computing $y_l$}  & \multicolumn{1}{c}{\scriptsize Computing $y_r$ }\\
 
%[-0.4em] 
%......................................................
%$\bunderline[3]{f}_j = 0, \forall j \in [1,M]$
\hline
\footnotesize \textbf{1}&\multicolumn{2}{c}{\scriptsize $\bunderline[3]{f}_j = 0, \forall j \in [1,M]$, then}\\   
\footnotesize &\multicolumn{2}{c}{\scriptsize $y_l$ = min($\bunderline[3]{w}_{j}$), \hspace{3cm}  $y_r$ = max($\oversymb{w}_{j}$) }\\   
\footnotesize &\multicolumn{2}{c}{\scriptsize $\forall j \in [1, M]$ with $\oversymb{f}_j \neq 0$, Stop}\\
\footnotesize \textbf{2}&\multicolumn{2}{c}{\footnotesize Initialise z$_j = 1$, $\Delta w_{j} = \oversymb{f}_j - \bunderline[3]{f}_j, \forall j \in [1,M]$  }\\ 
\footnotesize \textbf{3}&\multicolumn{2}{c}{\footnotesize Calculate: }\\
&\multicolumn{2}{c}{\footnotesize $\Bigg\{ \delta_1 = \displaystyle \sum_{j = 1}^M  \oversymb{f}_j, \delta_{2} = \sum_{j = 1}^M \oversymb{f}_j w_{j}$\Bigg\} }\\
\footnotesize \textbf{4}&\multicolumn{2}{c}{\footnotesize $flag = 0$ }\\
\footnotesize \textbf{5}&\multicolumn{2}{c}{\footnotesize For $j = 1$ to $M$, repeat the following operations of this step }\\
\footnotesize &\multicolumn{2}{c}{\footnotesize $A_j = x_j \delta_1 - \delta_2$ }\\
&\footnotesize If $A_j < 0$ & \multicolumn{1}{c}{\footnotesize If $A_j >0$ }\\ &\multicolumn{2}{c}{\scriptsize z$'_{j} =1$, else z$'_{j} =0$ } \\
&\multicolumn{2}{c}{\footnotesize If z$'_{j} \neq z_{j}$ then} \\
&\multicolumn{2}{c}{\footnotesize If z$_j = 1,$ $
    \begin{cases}
     flag = 1,~\text{z}_{1} = \text{z}_{1}  + \Delta w_{j} \\
 \text{z}_{j} = \text{z}'_j,~~~\text{z}_{2}=\text{z}_{2} + w_j \Delta w_{j}
    \end{cases}$ }  \\
&\multicolumn{2}{c}{\footnotesize Else $
    \begin{cases}
     flag = 1,~\text{z}_{1} = \text{z}_{1}  - \Delta w_{j} \\
 \text{z}_{j} = \text{z}'_j,~\text{z}_{2}=\text{z}_{2} - w_j \Delta w_{j}
    \end{cases}$ }  
\\
\footnotesize \textbf{6}&\multicolumn{2}{c}{\footnotesize If $flag \neq 0$ \footnotesize go to step \textbf{4}; else}\\
&\multicolumn{1}{c}{\footnotesize $y_l = \delta_2/\delta_1$; $z_{lj} = z_j$}& \multicolumn{1}{c}{\footnotesize $y_r = \delta_2/\delta_1$; $z_{rj} = z_j$}\\
\hline
%==================================================
		\end{tabular}
		\centering % used for centering table
		\label{table:SC_algorithm} % is used to refer this table in the text
\end{tablehere}
%++++++++++++++++++++++++++++++++++++++++++++++
\begin{equation}
\bunderline[3]{f}_j(\vec{x}_p) = \prod_{j=1}^M \bunderline[3]{\mu}_{jk},~~\oversymb{f}_j(\vec{x}_p) = \prod_{j=1}^M \oversymb{\mu}_{jk}
\end{equation}
In traditional IT2-FISs, an iterative Karnik-Mendel (KM) method is usually employed to calculate each type-reduced set ($y_l, y_r$) and the associated switching points \cite{chen2020comprehensive}. This information is used to calculate the output of the IT2-FIS \cite{rubio2014interval,rubio2016data}. To avoid the iterative nature of KM algorithms, the proposed FIT2-FELM is used as a fast training method for IT2 Fuzzy Inference Systems (IT2-FISs) with a simplified version of the Center of Set Type Reducer Without Sorting Requirement Algorithm (COSTRWSR) called SC method \cite{khanesar2016improving}. Based on the SC algorithm (see Table I), the reduced set $y_{l}$ and $y_{r}$ can be obtained \cite{khanesar2016improving}:

\begin{equation}
y_l = \frac{ \sum_{j = 1}^M \oversymb{f}_j w_{j} - \sum_{j=1}^M (1 - z_{lj})\Delta w_{j} w_{j} }{ \sum_{j = 1}^M \oversymb{f}_j - \sum_{j=1}^M (1 - z_{lj})\Delta w_{j}}
\end{equation}
\begin{equation}
y_r = \frac{ \sum_{j = 1}^M \oversymb{f}_j w_{j} - \sum_{j=1}^M (1 - z_{rj})\Delta w_{j} w_{j} }{ \sum_{j = 1}^M \oversymb{f}_j - \sum_{j=1}^M (1 - z_{rj})\Delta w_{j}}
\end{equation}
% in which:
% \begin{multline}
%     \varphi_{pj} = \frac{ \oversymb{f}_j - (1 - z_{lj})\Delta w_{j}  }{ \sum_{j = 1}^M \oversymb{f}_j - \sum_{j = 1}^M(1 - z_{lj})\Delta w_{j}}
%     + \\ \frac{ \oversymb{f}_j - (1 - z_{rj})\Delta w_{j} }{ \sum_{j = 1}^M \oversymb{f}_j - \sum_{j=1}^M (1 - z_{rj})\Delta w_{j}} 
% \end{multline}
where $\Delta w_{j} = \oversymb{f}_j - \bunderline[3]{f}_j, \forall j \in [1,M]$.
To the training of a TSK IT2-FIS and to determine each $q_{j,k}$, the proposed FIT2-FELM is implemented in two main steps:
\begin{itemize}
    \item \textbf{Step 1. Random initialisation} of each MF's parameter $m_{jk}$ and $\sigma_{jk}$ in the TSK IT2-FIS
    \item \textbf{Step 2. Calculation of each consequent} $q_{ij,k}$ by solving the linear system that results from the compact representation of Eq. (1) \cite{deng2013t2fela}:
\begin{equation}
    \textbf{H}\textbf{Q}_A =  \textbf{T}
\end{equation}
in which $\textbf{H} = [\textbf{h}_1, \ldots, \textbf{h}_P]^T$ is the matrix that contains all the firing strengths:
\begin{multline}
   \textbf{h}_p = \frac{1}{2}[\varphi_{p1} x_{p1}, \ldots ,  \varphi_{p1} x_{pN}, \ldots \\
   \varphi_{pM} x_{p1}, \ldots ,\varphi_{pM} x_{pN}]
\end{multline}
        where $\textbf{h}_p \in \textbf{R}^{1 \times (M \times N)}$ and $\textbf{T} = [\textbf{t}_1, \ldots, \textbf{t}_P]^T$ is the target sonar data, and each term $\varphi_{pj}$ is obtained by combining Eq. (6) and (7):
        \begin{multline}
    \varphi_{pj} = \frac{ \oversymb{f}_j - (1 - z_{lj})\Delta w_{j}  }{ \sum_{j = 1}^M \oversymb{f}_j - \sum_{j = 1}^M(1 - z_{lj})\Delta w_{j}}
    + \\ \frac{ \oversymb{f}_j - (1 - z_{rj})\Delta w_{j} }{ \sum_{j = 1}^M \oversymb{f}_j - \sum_{j=1}^M (1 - z_{rj})\Delta w_{j}} 
\end{multline}
in which the terms $[z_{lj},z_{rj}]$ are obtained by using the SC algorithm.
$\textbf{Q}_A$ is the consequent matrix $\textbf{Q}_A=[\textbf{c}_1, \ldots, \textbf{c}_M]^T$, such as $\textbf{c}_j = [q_{j,1}, \ldots, q_{j,N}]$. Thus, based on the constraint of minimum norm least square and ELM theory, the training of a TSK IT2-FIS can be performed by solving the linear system in Eq. (8)
\begin{equation}
    \stackengine{4pt}{\text{min}}{$x$}{U}{c}{F}{T}{S} ||\textbf{Q}_A|| \text{~and~}\stackengine{4pt}{\text{min}}{$x$}{U}{c}{F}{T}{S} ||\textbf{HQ}_A - \textbf{T}||
\end{equation}
\end{itemize}
Hence, the optimal solution of Eq. (8) is $\textbf{Q}_A = \textbf{H}^{\dagger} \textbf{T}$, such as $H^{\dagger}$ is the Moore-Penrose generalised inverse of $\textbf{H}$ \cite{deng2013t2fela}.
%++++++++++++++++++++++++++++++++++++++++++
\begin{figurehere}
\begin{center}
%\centering
%\vspace{2mm}
\includegraphics[width=8.5cm, height=7.5cm ]{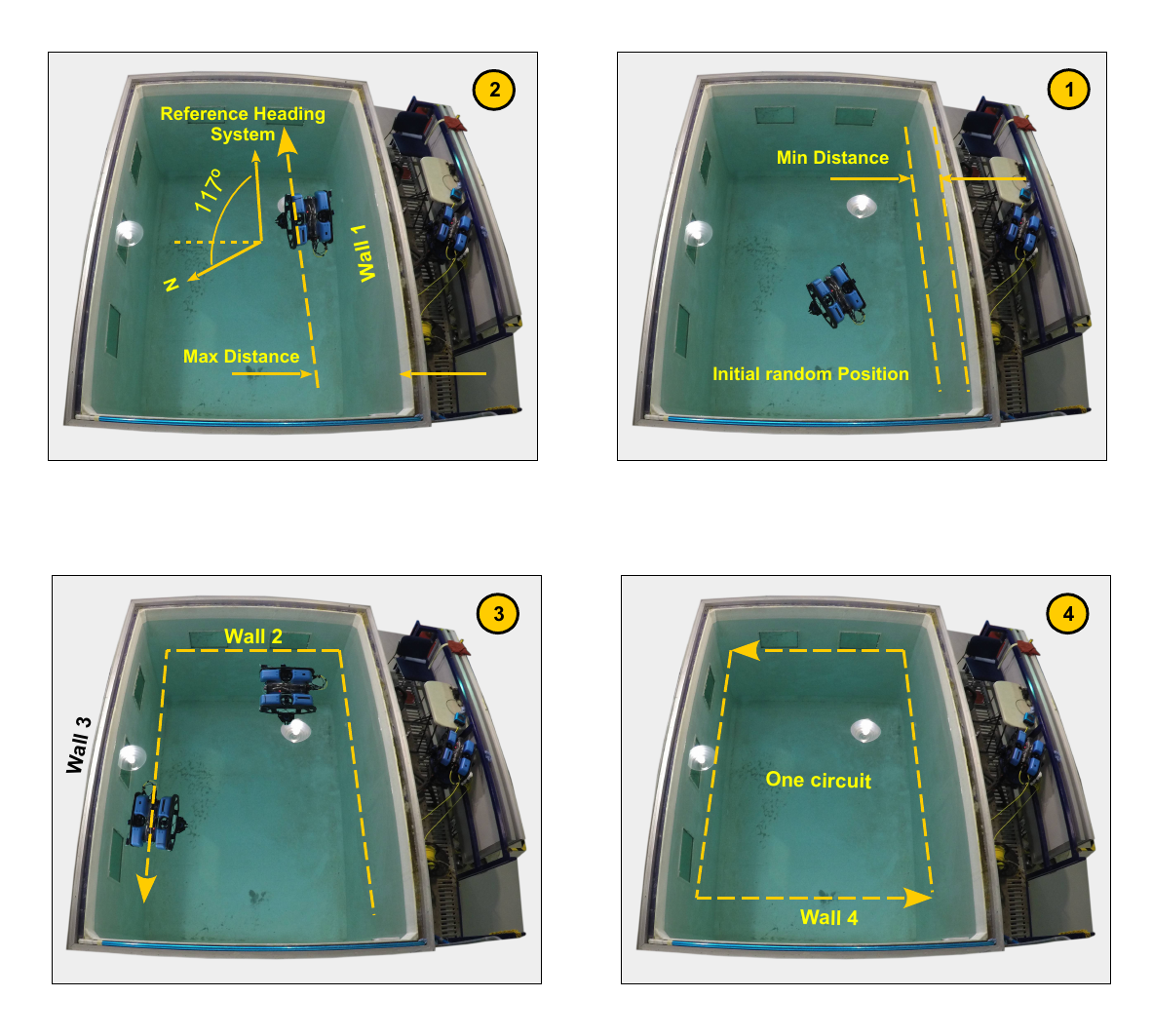}
\caption{One circuit completed by the BlueROV2.}\label{fig::one_circuit}
\end{center}
\end{figurehere}
%++++++++++++++++++++++++++++++++++++++++++
%++++++++++++++++++++++++++++++++++++++++++
\subsection{On-board data classification for Path Following using a TSK IT2-FIS}
The above FIT2-FELM approach was used to training of a  TSK IT2-FIS for the on-board classification of sonar data in the BlueROV2. The IT2-FIS was integrated into a Hierarchical Navigation Strategy (HNS) as a guidance control to provide the BlueROV2 with full autonomy necessary to navigate predefined missions.

Based on the structure of the water container used for all the experiments, one circuit is completed by the BlueROV2 following a clockwise/anticlockwise path navigating along walls 1-2-3-4/4-3-2-1 correspondingly (see Fig. \ref{fig::one_circuit}). For the successful completion of one circuit, the HNS performs two hierarchical levels of control. First, at low-level control, the HNS decomposes the control motion into four simple fuzzy local behaviours, i.e. 1) heading behaviour, 2) depth behaviour, 3) edge-distance behaviour, and 4) contour classification based on an IT2-FIS. The first three behaviours involve the application of Interval Type-2 Fuzzy Proportional Derivative Controllers (IT2-FPDCs) that allow the BlueROV2 to maintain a parallel alignment with respect to each wall at a predefined depth and distance respectively. In the fourth behaviour, the BlueROV2 employs the IT2-FIS to identify the contour around it, namely, a wall or a corner. Thus, a binary classification is performed by the TSK IT2-FIS in which its output $y_p = \{wall=0, corner=1\}$, and five angle readings provided by the sonar data are used as inputs $\textbf{x}_p = [180^o, 172^o, 164^o, 156^o, 148^o]$. From the collected data, a number of $400$ and $388$ records corresponds to class $0$ and $1$, respectively. Coordination of low-level behaviours is achieved by a high-level hierarchy that follows the principle operation used by fuzzy architectures.  At this level, the BlueROV2's motion is encoded as a base with $2^3$ rules that maps relevant sensor inputs to control outputs according to a desired control policy. Activation of each behaviour follows is based on a $qth$ fuzzy rule with two premises \cite{rubio2016iterative}:
\begin{multline}
R^q: \text{Yaw angle}~\text{is}~\text{large ($\beta_{cp}\geq \beta_{ref} + 5^o$)} ~\text{AND}~ \ldots 
\\
\text{ROV's depth}~\text{is}~\text{small ($d_{ROV}\leq d_{ref}$)}~\ldots
\\
~\text{THEN}~\text{depth and heading behaviour}
\end{multline}
Each IT2-FPDC is configured with three triangular Membership Functions (MFs) and a Wu-Mendel type-reducer, in which error ($e$) and its derivative ($\Delta e$) are the inputs to the fuzzy controller.
%+++++++++++++++++++++++++++++++++++++
%++++++++++++++++++++++++++++++++++++++++++
Pulse Width Modulation (PWM) using an IT2 FPDC is applied to control the angular speed of each thruster: 
\begin{equation}
    \begin{aligned}
       E(k) = G_e e(k) = G_e(y_{ref} - y_{f})\\
       \Delta E(k) = G_{\Delta e} \Delta e(k) = G_{\Delta e}(e(k) - e(k-1))
    \end{aligned}
\end{equation} 
where $y_{ref}$ and $y_f$ is the reference signal and output system correspondingly while $k$ is the sampling index. 
\begin{figurehere}
\begin{center}
%\centering
%\vspace{2mm}
\includegraphics[width=9cm, height=8cm ]{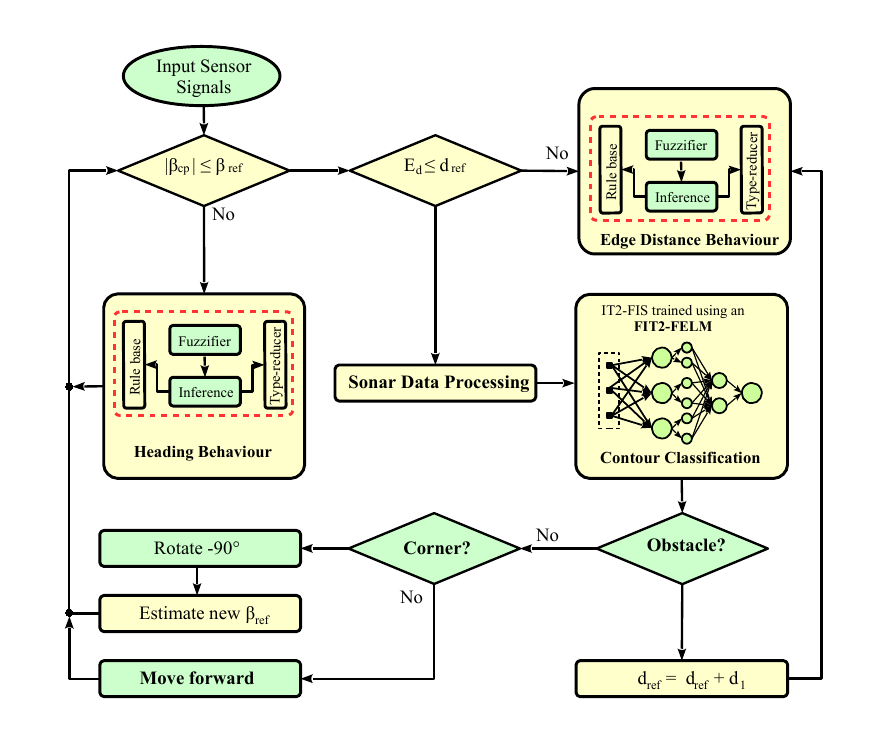}
\caption{\footnotesize Architecture of the HNS at low-level hierarchy.}\label{fig::HNS}
\end{center}
\end{figurehere}

The output variable $u(k)$ is calculated from the incremental output $\Delta u(k)$ and its previous value $u(k-1)$ by: 
\begin{equation}
    u(k) = u(k-1) + \Delta (k)
\end{equation}
in which, $\Delta u(k) = G_{U} U(k)$, in which, $G_{U}$ is an scaling factor. As illustrated in Fig. \ref{fig::one_circuit}, to begin with each experiment (path), the BlueROV2 is randomly placed in the container. From there, to implement the heading behaviour, the reference angle $\beta = 0$, the target heading $\beta_{ref}$ is parallel to the nearest wall $\beta_{cp} = \beta_{ref}$ ( Control inputs, $e = \beta_{ref} - \beta_{cp}$,  and its change $\Delta e = e(k) - e(k-1)$). The angle $\beta_{ref}$ can be calculated by adding the current heading $\beta_{cp}$ and $\theta_m$ that are collected by an onboard compass and ping sonar respectively.
\begin{equation}
    \beta_{ref} = \beta_{cp} + \theta_m
\end{equation}
To implement the edge distance behaviour, $d_w$ is the target distance to the closest wall and $d_c$ is a single distance beam provided by the ping sonar (mts).
\begin{tablehere}
\captionsetup{font=footnotesize, justification=centering}
\caption{ Average performance for five-cross-validation and on-board classification.}\label{iris_results} % title of Table
%\vspace{0.0cm}
\centering % used for centering table
\begin{tabular}{p{2.00cm} |p{0.1cm}|p{0.1cm} |p{0.9cm}|p{0.1cm} |p{0.1cm}| p{0.1cm} | p{1.5cm} }
\hline 
%inserts double horizontal lines
 %& \multicolumn{2}{c}{RMSE} \\
 \scriptsize Model &\multicolumn{3}{c}{\scriptsize Training ($\%$)}  & \multicolumn{3}{c}{\scriptsize Testing ($\%$)} & \scriptsize On-board Classification\\
%[-0.4em] 
%......................................................
 
\hline
& \multicolumn{2}{c}{\scriptsize Average}&\scriptsize Time (s)&\multicolumn{3}{c}{\scriptsize Average}&\multicolumn{1}{c}{ \scriptsize $\%$}\\   

\hline 
%...................................................... 
% \multicolumn{8}{c}{\scriptsize  \textbf{MNIST} } \\
% \hline
  \scriptsize  FELM & \multicolumn{2}{c}{\scriptsize 95.28} & \multicolumn{1}{c}{\scriptsize 3.22} &\multicolumn{3}{c}{\scriptsize 94.19} & \multicolumn{1}{c}{\scriptsize 88.12 }\\
 \scriptsize  EGT2-FELM& \multicolumn{2}{c}{\scriptsize 99.31} & \multicolumn{1}{c}{\scriptsize 540.2} &\multicolumn{3}{c}{\scriptsize 98.73} & \multicolumn{1}{c}{\scriptsize  93.79}\\
  \scriptsize  FELM-IT2FELM& \multicolumn{2}{c}{\scriptsize 97.79} & \multicolumn{1}{c}{\scriptsize 3.34} &\multicolumn{3}{c}{\scriptsize 97.42}&\multicolumn{1}{c}{\scriptsize  92.75}\\
  \scriptsize IT2-FELM-KM  & \multicolumn{2}{c}{\scriptsize 97.67} & \multicolumn{1}{c}{\scriptsize 12.29} &\multicolumn{3}{c}{\scriptsize 97.40}&\multicolumn{1}{c}{\scriptsize  92.71}\\
  \scriptsize ELM  & \multicolumn{2}{c}{\scriptsize 90.21} & \multicolumn{1}{c}{\scriptsize 0.93} &\multicolumn{3}{c}{\scriptsize 86.22}&\multicolumn{1}{c}{\scriptsize  78.11}\\
\hline
%====================================================
%**************************************************
	%	[1ex] % [1ex] adds vertical space
		
		\end{tabular}
		\centering % used for centering table
		\label{nfm_results} % is used to refer this table in the text
\end{tablehere}
%++++++++++++++++++++++++++++++++++++++++++
%+++++++++++++++++++++++++++++++++++++++++++++
\begin{figurehere}
\begin{center}
\includegraphics[width=8cm, height=6.2cm ]{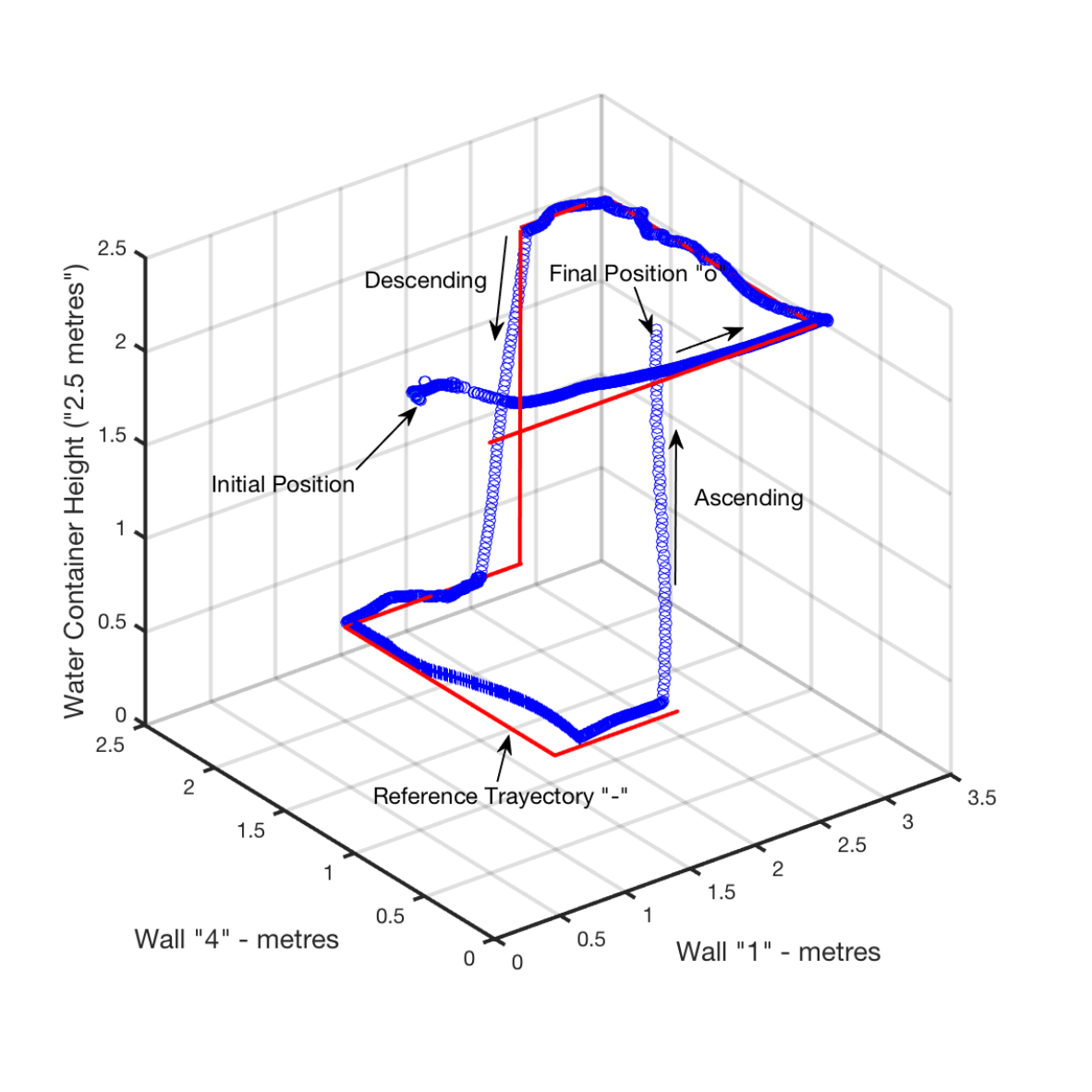}
\caption{\footnotesize Target (red line "$-$") and current trajectory (blue line, "$o$") followed by the BlueROV2 at two different depths of a random experiment. }\label{fig::trajectory}
\end{center}
\end{figurehere}
To control the vertical position of the BlueROV2, an IT2-FPD with two inputs, namely, $e$ and $\Delta e$ is applied, where $e = d_{ref} - d_{ROV}$. $d_{ref}$ is the target depth, $d_{ROV}$ is the current vertical position provided by a depth sensor in $cm$ and $\Delta e = e(k) - e(k-1)$.

\section{Results}
To verify the efficiency of an FIT2-FELM, two different experiments are suggested. First, the proposed FIT2-FELM is applied to the trainining of a TSK IT2-FIS with a SC type-reduction layer and six fuzzy rules. A sonar data set of $788 \times 5$ records collected offline was employed for five-cross-validation. The sonar data set contains $420$ and $368$  records for class wall and corner respectively.

As detailed in the first column of Table II, a number of existing learning methodologies was applied to the training of a TSK IT2-FIS, i.e. using an Evolutionary General Type-2 Fuzzy ELM (EGT2-FELM), an IT2 Fuzzy ELM with a Karnik-Mendel algorithm (IT2-FELM-KM),a Fuzzy ELM (FELM) of type-1 and the traditional ELM. Evidently, by including a SC algorithm into the IT2-FIS structure improves its trade-off between computational load and model accuracy. As illustrated in Fig. \ref{fig::trajectory}, the second type of experiment involves the completion of a path of two circuits at two distinct depths, i.e. $d_{ref} = 0.0$ metres and $d_{ref} = 2.0$ metres. In this experiment, on-board sonar data classification is performed by implementing the trained TSK IT2-FIS in the Raspberry Pi3. Parameters for this experiment include a target edge distance $d_w = 0.65 ~\text{mtrs}$ and arbitrarily set to $d_w = 1.5 ~\text{mtrs}$ along wall 4. As can be observed from Fig. \ref{fig::trajectory}, the BLUEROV2 is randomly located at the beginning of the first circuit. To control the behaviours edge distance, heading behaviour and depth control, an IT2-FPDC with three triangular fuzzy rules and a Wu-Mendel type-reducer \cite{wu2012overview} were implemented.

%++++++++++++++++++++++++++++++++++++++++++++++
\begin{figurehere}
\begin{center}
\includegraphics[width=9.3cm, height=10cm ]{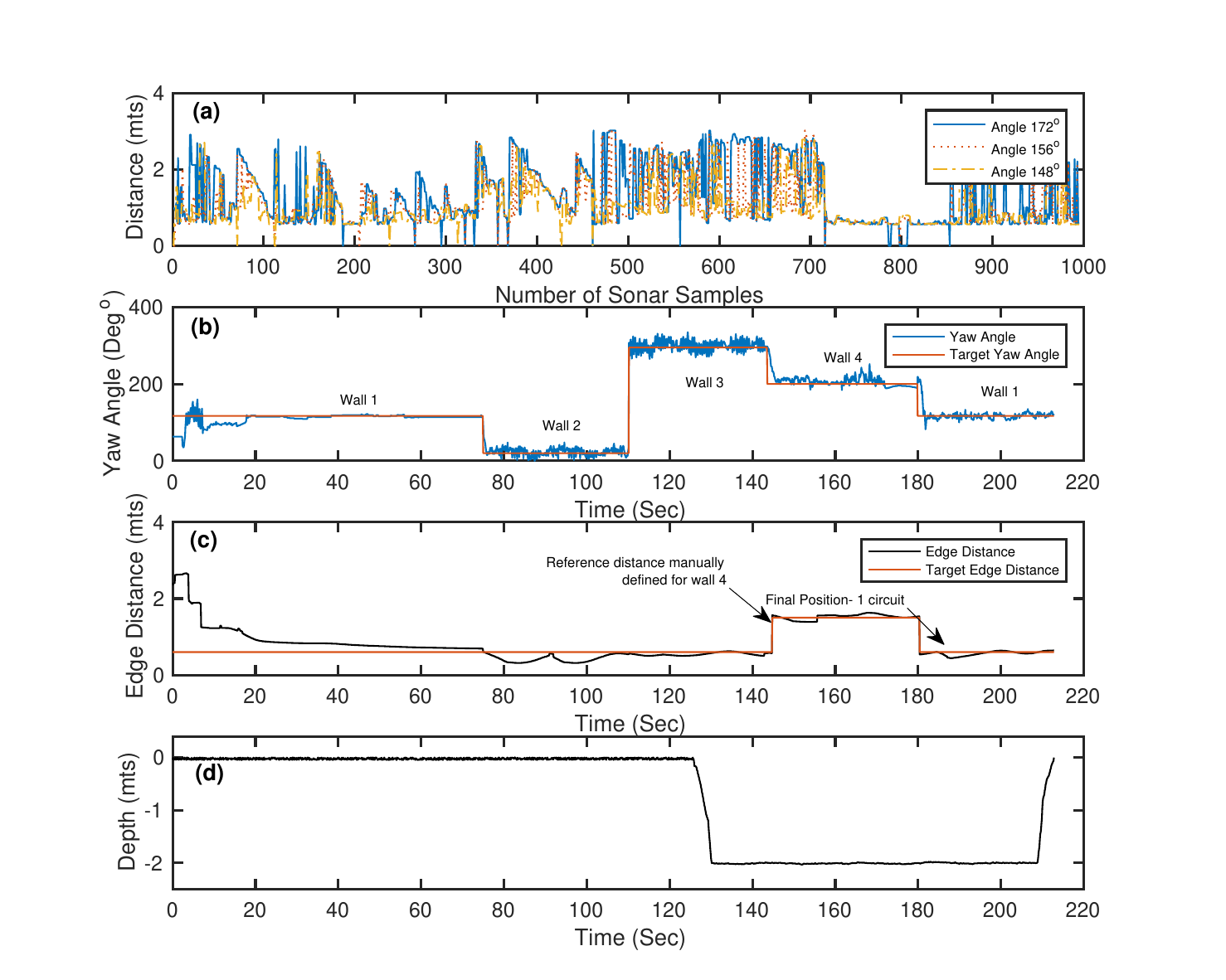}
\caption{\footnotesize (a) Sample Sonar Signals for angles $[172^o, 156^o, 148^o]$ and (b-d) BlueROV2's position of a random experiment.}\label{fig::fuzzy_behaviours_Control}
\end{center}
\end{figurehere}
$x_k$ is the input sonar data collected at time $k$, and the term $c$ denotes the current class, . Therefore, $y_k$ is computed by:
\begin{equation}
    y_k(c|x_k) = \frac{1}{n_s}\sum_{k=k-n_s}^{k} y_{k}(x_k);~n_s = 4
\end{equation}
In Fig. \ref{fig::fuzzy_behaviours_Control}(a), sonar data samples that correspond to the angles $[172^o, 156^o, 148^o]$ is illustrated. The BlueROV2's position for a random circuit is also presented in Fig.  \ref{fig::fuzzy_behaviours_Control}(b-d).   
\begin{figurehere}
\begin{center}
%\centering
%\vspace{2mm}
\includegraphics[width=6cm, height=5.5cm ]{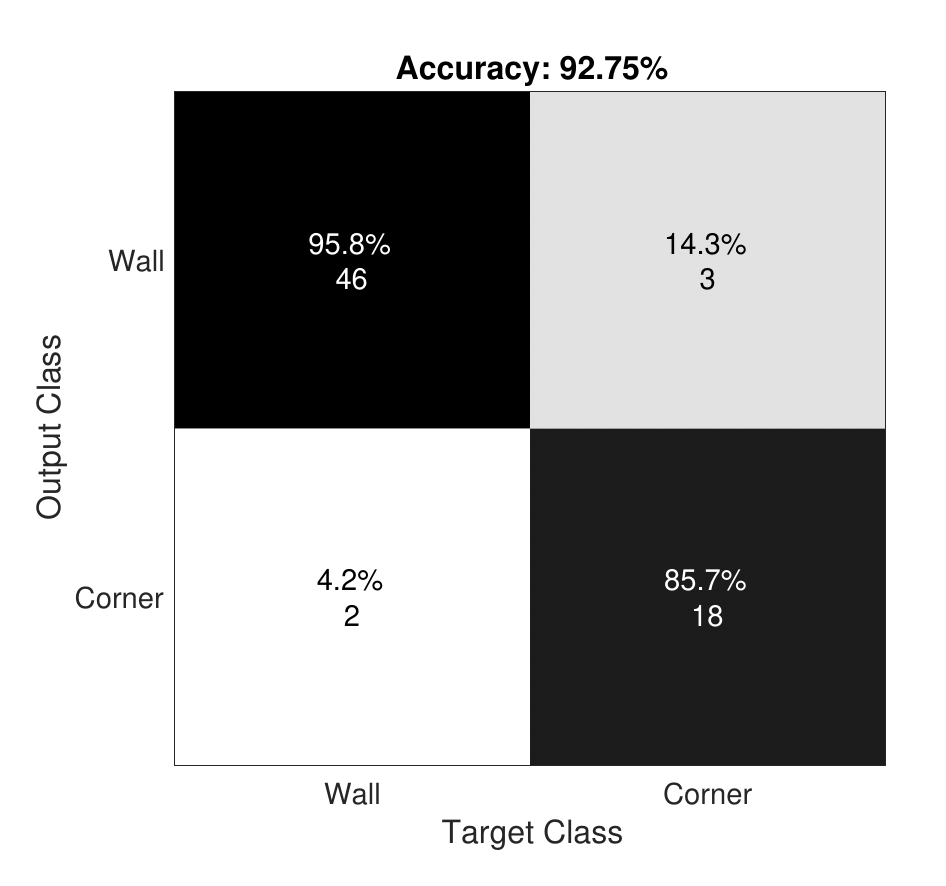}
\caption{\footnotesize Average Confusion Matrix of five random experiments for on-board sonar data classification using a TSK IT2-FIS trained with FIT2-FELM.}\label{fig::Confusion_matrix}
\end{center}
\end{figurehere}
\begin{figurehere}
\begin{center}
%\centering
%\vspace{2mm}
\includegraphics[width=9.5cm, height=4cm ]{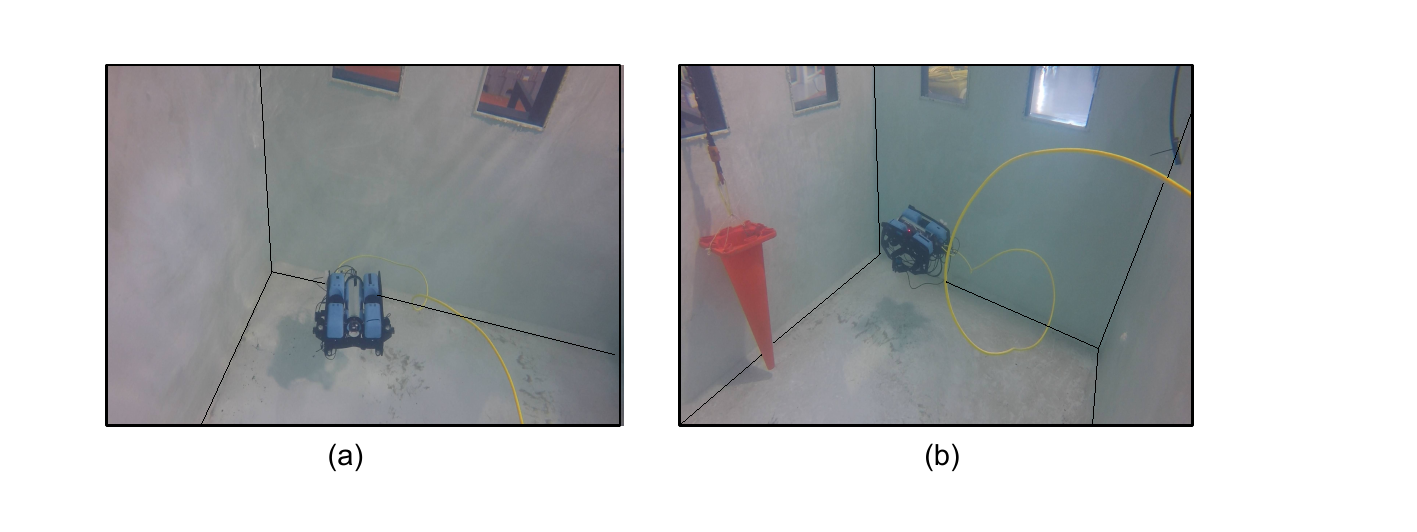}
\caption{\footnotesize (a) BlueROV2 perceiving a corner and (b) autonomously navigating at a depth of two metres in the presence of an obstacle.}\label{f ig::BlueROV_exploring}
\end{center}
\end{figurehere}
The corresponding confusion matrix of the average of five experiments is presented In Fig. \ref{fig::Confusion_matrix}. As observed in the matrix, under real world conditions only a few sonar data is available to classify each corner (see Fig. \ref{f ig::BlueROV_exploring}).  
We also observed from the experiments, it can be observed identifying corners produces the highest value fin the confusion matrix, particularly when target edge distance is set $d_w = 1.5~mtrs$.  

In summary, the classification accuracy was good enough to perceive corners and walls while autonomously completing a predefined path at two different depths. By using FIT2-FELM, a high balance between the average training time, low computational load for real sonar data classification, model simplicity and accuracy was achieved. 
%++++++++++++++++++++++++++++++++++++++++++
\section{Conclusions and Future Work}

For this article, a Fast Interval Type-2 Fuzzy Extreme Learning Machine (FIT2-FELM) to train a class of IT2 Fuzzy Inference Systems of Takagi Sugeno Kang (TSK IT2-FISs) for on-board sonar data classification was suggested. The trained IT2-FIS was implemented into a Hierarchical Navigation Strategy (HNS) as a guidance mechanism that provides an underwater vehicle  called BlueROV2 with full autonomy to complete predefined paths. For local planning, the HNS decomposes the BlueROV2's motion into a bottom-up hierarchy of high and low level behaviours. At the low level, the HNS breaks down the navigation strategy into a number of four local fuzzy behaviours i.e. 1) heading behaviour, 2) depth behaviour, 3) edge distance and 4) contour classification. At the high level, the BlueROV2's motion is encoded as a fuzzy rule base that maps relevant sensor inputs into a desired planning goal.  

To evaluate the proposed approach, a number of experiments to the completion of a predefined path at two depths in a water container of $2.5m \times 2.5m \times 3.5m$ was suggested. From experiments, we can observe that local motions are treated by the HNS as building blocks for more intelligent composite behaviours. The successful completion of one circuit implies that goal-directed navigation is decomposed by the HNS into a behavioral mechanism for wall-following, contour classification and local planning. In essence, progressing to a more complex navigation strategy mainly depends on the correct information produced by the IT2-FIS. Therefore, providing the BlueROV2 with a complete picture of its surroundings mainly depends on the classification outcome and sonar information.

Using FIT2-FELM to train a TSK IT2-FIS not only accounted to a better treatment of noisy sonar data, but also to reduce the associated computational load that results from the use of IT2 Fuzzy Sets (FSs). 

Experiments presented in this study focus on autonomous navigation in controlled underwater conditions. The proposed robotic system can also be used in real-world conditions. Upcoming research involves autonomous navigation in open water environments.

%%%%%%%%% REFERENCES
{\small
\bibliographystyle{ieee_fullname}
\bibliography{egbib}
}

\end{document}